\documentclass[11pt,a4paper]{article}
\usepackage[hyperref]{emnlp2020}
\usepackage{times}
\usepackage{latexsym}

\usepackage{placeins}
\usepackage{microtype}
\usepackage{booktabs}
\usepackage{multirow}
\usepackage{graphicx}
\usepackage{comment} 
\usepackage{mathtools}
\usepackage{todonotes}
\usepackage{makecell}
\usepackage{subcaption}

\usepackage[linesnumbered,boxed]{algorithm2e}
\usepackage[noend]{algpseudocode}

\usepackage{amsmath,amsfonts,bm}

\def\eqref#1{equation~\ref{#1}}

\def\1{\bm{1}}

\def\vk{{\bm{k}}}

\def\vo{{\bm{o}}}

\def\vq{{\bm{q}}}

\def\vv{{\bm{v}}}

\def\vx{{\bm{x}}}

\DeclareMathAlphabet{\mathsfit}{\encodingdefault}{\sfdefault}{m}{sl}
\SetMathAlphabet{\mathsfit}{bold}{\encodingdefault}{\sfdefault}{bx}{n}

\usepackage{microtype}

\aclfinalcopy 

\title{Can Transformers \textit{Jump Around Right} in Natural Language? \\ Assessing Performance Transfer from SCAN}

\author{%
Rahma Chaabouni$^{* \dagger}$ \\
Ecole Normale Superieure \\
  \texttt{\hspace{11cm}\{chaabounirahma, roberto.dessi11\}@gmail.com kharitonov@fb.com}
  \And 
Roberto Dess{\`\i} \\
Facebook AI \& Pompeu Fabra \And
Eugene Kharitonov\thanks{\hspace{.15cm} Equal contribution. $^\dagger$ work was done while R.C. was at Facebook AI.} \\
Facebook AI
}

\date{}

\begin{document}
\maketitle
\begin{abstract}
Despite their failure to solve the compositional SCAN dataset, seq2seq architectures still achieve astonishing success on more practical tasks. This observation pushes us to question the usefulness of SCAN-style compositional generalization in realistic NLP tasks.  
In this work, we study the benefit that such compositionality brings about to several machine translation tasks. We present several focused modifications of Transformer that greatly improve generalization capabilities on SCAN and select one that remains on par with a vanilla Transformer on a standard machine translation (MT) task. Next, we study its performance in low-resource settings and on a newly introduced distribution-shifted English-French translation task. 

Overall, we find that improvements of a SCAN-capable model do not directly transfer to the resource-rich MT setup. 
In contrast, in the low-resource setup, general modifications lead to an improvement of up to 13.1\% BLEU score w.r.t.\ a vanilla Transformer. Similarly, an improvement of 14\% in an accuracy-based metric is achieved in the introduced compositional English-French translation task.
This provides experimental evidence that the compositional generalization assessed in SCAN is particularly useful in resource-starved and distribution-shifted scenarios.

\end{abstract}
\section{Introduction}
\label{s:intro}
While sequence-to-sequence (seq2seq) models achieve remarkable performance in many tasks~\citep{Sutskever2014,Raffel2019,Adiwardana2020}, they often fail to generalize in a systematic way~\cite{Baroni2019,Mccoy2020,Hupkes2020,Kharitonov2021,Dankers2021}. These shortcomings are particularly obvious in the experiments on the SCAN domain~\cite{Lake2018,Loula2018,Bastings2018}. 

In SCAN, 
inputs are instructions that describe trajectories and outputs define sequences of actions to follow them (see Table~\ref{fig:scan-examples}). 
To illustrate how SCAN probes a model for compositional generalization, imagine that we train it on a set of instructions \{\texttt{jump}, \texttt{run}, \texttt{turn}, \texttt{turn twice}, \texttt{run twice}\}, but test it on \texttt{jump twice}. Strictly speaking, nothing in the training data indicates that the model \textit{must} output \texttt{JUMP JUMP} instead of e.g.\ \texttt{JUMP}. However, it is hypothesised that a bias for such compositional,  human-like induction is beneficial~\cite{Lake2018,Lake2019}.


This hypothesis stumbles into a perplexing situation: despite failing at compositional generalization, considered as a core requirement for language understanding, seq2seq models have tremendous success in practice. 
Is the type of compositional generalization, that SCAN probes for, useful for NLP tasks? If so, in what scenarios?


In this work, we aim to answer this question. 
Firstly, we introduce focused modifications to Transformer 
that greatly improve accuracy performance on SCAN. 
To build such modifications, we exploit two observations: (i) Transformer's architecture is very similar to convolution-based seq2seq models (ConvS2S)~\citep{Gehring2017}, (ii) ConvS2S performs well on SCAN~\citep{Dessi2019}. This capability (ii) is hypothesized to be due to explicitly localized representation of the sequences, where only deep layers of the model can access more distant tokens~\cite{Dessi2019,Hupkes2020}.  Such a capability should also benefit to natural language  processing as human languages are proved to favor local syntactic constructions~\cite{Futrell2015}. Motivated by these observations, we focus on the major differences between ConvS2S and Transformer, namely the localized receptive field span and the gating mechanism, to inject into Transformer inductive biases useful for SCAN.

As a testbed,  
we use the machine translation (MT) domain as one of the most popular applications for seq2seq models. We consider both resource-rich (IWSLT’14 German$\rightarrow$English) and low-resource (Nepali\&Sinhala$\leftrightarrow$English) setups. Finally, to evaluate SCAN-style capabilities in natural language tasks, we build a dataset that probes whether models can systematically generalize w.r.t.\ noun-adjective ordering while translating from English to French. We construct this dataset by filtering the EuroParl corpus (a part of WMT'14).



Our results indicate that combining two ConvS2S-inspired changes
improves accuracy on one SCAN split (SCAN-jump) from 3.4\% to 43.0\%, while maintaining a high accuracy on the other splits (SCAN-simple and SCAN-around-right).
As expected, given that SCAN is an artificial diagnostic dataset, 
not all modifications lead to equal 
improvements on an MT task. We select one of the considered modifications that performs on par with the vanilla Transformer on IWSLT'14.

Testing the selected modification on low-resource data, we observe that it provides between 3.6\% and 13.1\% BLEU improvements over Transformer. On the noun-adjective ordering dataset, we find that our modification results into gains in generalization of 14\%. 

This leads to the following picture: the localized attention, augmented by a gating mechanism, provides a useful inductive bias that proves to be beneficial for SCAN-style generalization. 
Additionally, it turns out useful in low-resource and distribution-shifted settings. Thus, testing seq2seq models on SCAN while controlling for a non-degraded performance leads to improvement in domains where syntactic compositionality is crucial for a task success.

\begin{table*}[t]
  \centering
  \resizebox{0.9\textwidth}{!}{
  \begin{tabular}{lll}
    jump&$\Rightarrow$&JUMP\\
    jump around right&$\Rightarrow$&RTURN JUMP RTURN JUMP RTURN JUMP RTURN JUMP\\
    turn left twice&$\Rightarrow$&LTURN LTURN\\
     jump opposite left after walk around left&$\Rightarrow$&LTURN WALK LTURN WALK LTURN WALK LTURN WALK LTURN LTURN JUMP\\
  \end{tabular}
  }
  \caption{Examples of SCAN trajectories and instructions, adopted from \citep{Lake2018}.}
  \label{fig:scan-examples}
\end{table*}

\section{Transformers and ConvS2S}
\label{s:archs}

\textbf{Architecture overview} Both Transformer and ConvS2S are encoder-decoder architectures~\cite{Sutskever2014}, where the decoder has an attention mechanism to peek into the encoder's representation~\cite{Bahdanau2014}. These representations are obtained by embedding the inputs, adding a positional embedding, and passing them through a sequence of layers. 


In Transformer's encoder, the output representations are the result of a sequential application of two (sub)layer types: self-attention and fully connected layers. The input representation can ``skip'' any sublayer via a residual connection. The output of the sublayer is passed through a dropout mechanism and added to the residual. This sum is then  layer-normalized. 
Any relation between input tokens is modeled solely by self-attention modules.

In ConvS2S, the encoder is also a sequence of identical blocks. The inter-term dependencies are modeled by 1D convolutions with GLU activation functions~\cite{Dauphin2017}.\footnote{GLU was introduced as a combination of a convolution network and an activation; we follow the Pytorch convention and consider it as separate blocks for convenience.} In contrast to self-attention, convolutions have a finite kernel size, thus effectively capping the maximal distance of intermediate dependencies that can be modeled. The GLU activation function serves as a gate, allowing ConvS2S to control the balance between residuals and the output of the convolution. After the GLU operation, the intermediate representation is added to the residual and scaled. The output of the final convolution is then passed into a fully connected layer.
In ConvS2S and Transformer, decoders have similar structures to those of encoders, 
with an additional decoder$\rightarrow$encoder attention layer after the convolution and self-attention blocks, respectively. 

Despite the similarities, there are numerous low-level differences between the two architectures: normalization (layer norm~\cite{Ba2016} vs.\ weight normalization~\cite{Salimans2016}), optimization (Adam~\cite{Kingma2014} with a ramp-up vs.\ NAG~\cite{Sutskever2013}), etc. \textit{A priori}, any of those can affect models' inductive biases. However, we concentrate on some of the most obvious architectural differences: the limited convolution span and GLU activations. 

We believe these features can greatly affect models' performance on SCAN. Indeed, SCAN has only local dependencies between tokens, thus the ability to avoid spurious correlations with more distant tokens can be useful. Similarly, the ability to weight contributions from the token interactions into the intermediate representation is intuitively prerequisite to build compositional representation.


\textbf{GLU}~\cite{Dauphin2017} Given a vector input $\vx$, GLU splits it in two equally sized halves $\vx_1$ and $\vx_2$; one is passed through a sigmoid ($\sigma(x) = (1 + e^{-x})^{-1}$). Then both parts are pointwise multiplied:
\begin{equation}
\label{eq:glu}
    GLU(\vx) \coloneqq \sigma(\vx_1) \odot \vx_2
\end{equation}
This allows a network to implement a gating mechanism, where one half of its output gates the signal from the second. 

\textbf{Self-attention} Instead of convolutions, Transformer uses multi-headed self-attention to model interactions between tokens. Given $n$ embeddings, $\vx_{1}, \vx_2, ... \vx_n$ of dimensionality $d$, the self-attention layer transforms them in the following way.

Firstly, each embedding $\vx_i$ is projected by three matrices $Q$, $K$, and $V$ to get query $\vq_i$, key $\vk_i$, and value $\vv_i$ representations, respectively: $\vq_i, \vk_i, \vv_i \gets Q\vx_i, K\vx_i, V\vx_i$. Next, a scaled dot-product between query $\vq_i$ and key $\vk_j$ is calculated as follow:
\begin{equation}
\label{eq:self-attention}
    \alpha_{ij} = \frac{1}{\sqrt{d}} \vq_i  \cdot \vk_j^T
\end{equation}
This dot-product defines the attention weights $w_{ij} = e^{\alpha_{ij}} / \sum e^{\alpha_{ij}}$ which are used to get the output representations:
$\vo_i = \sum_j w_{ij} \vv_j$.
This process is done in parallel for multiple heads, acting on independent slices of the input embeddings; their outputs are concatenated and passed through a fully connected layer.


\section{Transformer Modifications}
\label{s:modifications}

\textbf{Self-attention gate (SAG)} The simplest way to imitate the effect of GLU activation (Eq.~\ref{eq:glu}) is to weight (gate) the output of self-attention by a learned scalar parameter. To ensure that it is non-negative and is scaled in $[0,1]$, we parameterize it as a sigmoid of a real-valued learned parameter $\beta$. Algorithm~\ref{fig:sag} illustrates the introduced change. In comparison to Transformer, SAG adds one scalar parameter for each encoder and decoder layer.

We treat $\beta_0$, the value $\beta$ is initialized with before training, as a hyperparameter. In the preliminary experiments, we found that after training, encoder layers often have small negative $\beta$ values ($-2..-0.5$), while decoder layers have positive values ($0.2..4.5$) that grow monotonically for higher layers. 

A similar modification was considered in an effort to stabilize Transformer training in the Reinforcement Learning domain~\cite{Parisotto2020}.

\textbf{Convolution as self-attention}
In the limit case, we can entirely replace self-attention with convolutions. This modification introduces one hyperparameter (kernel size). However, convolutional layers have fewer parameters than the self-attention mechanism.  One might consider this not to be a Transformer variant due to the lack of self-attention, but 
as self-attention generalizes convolutions~\cite{Cordonnier2020}, we consider this as an extreme form of regularization. 

\textbf{Fixed-span self-attention} A less extreme modification would be to use the regular multi-head self-attention mechanism, but without allowing attention to peek beyond some distance. This mimics the limited kernel size of convolutions in ConvS2S. We achieve this by adding a fixed bias term $b_{ij}$ to the self-attention logits (Eq.~\ref{eq:self-attention}): 
\begin{equation}
\label{eq:biased}
    \alpha_{ij} = \frac{1}{\sqrt{d}}\vq_i \cdot \vk_j^T + b_{ij}
\end{equation}
Setting $b_{ij}$ to $-\infty$ when the difference $|i - j|$ exceeds some fixed value $s$ and to 0 otherwise prevents the self-attention to look beyond distance $s$.


Fixed-span self-attention with a span parameter $s$  has the same ``receptive field'' as  1D convolution with kernel size $2s + 1$. This modification adds one hyperparameter (span size), but does not introduce new learned parameters.

\textbf{T5 attention} Further relaxing constraints on self-attention, we consider the case where we allow Transformer to learn how to (soft-)limit its self-attention. 
We introduce the bias term $b_{ij}$ that is learned as a function of a (signed) difference $i - j$, capping it to $[-s, +s]$ (e.g.,  positions with difference above $s$ would have the same bias $b_{s}$).  

This modification is similar to one introduced by \citet{Raffel2019} in T5, with the only exception that we allow each head to have its own bias. Again, the span size is a new hyperparameter. In a model with $n_h$ heads and $n_l$ layers, this modification requires $(2s + 1) \times n_l \times n_h$ new parameters, which is negligible in comparison with the sizes of fully connected layers. Examples of the learned $b_{ij}$ parameters are shown in Supplementary when training on SCAN.

\textbf{Implementation details} We used the fairseq~\cite{Ott2019} implementation of Transformer seq2seq as a foundation, with its initialization and default parameters. T5 and fixed-span attentions are implemented by providing additive attention masks to Transformer. 

\begin{algorithm}[tb]
\begin{algorithmic}[1]
\Procedure{compute\_SelfAttention}{}
\State $res \gets x$
\State $x \gets self\_attn(x)$
\State $x \gets x * \sigma (\beta)$
\State $x \gets layer\_norm(res + dropout(x))$
\EndProcedure
\end{algorithmic}
\caption{Self-attention gate (SAG). The only introduced change is on line 4. $\beta$ is a learned per-layer scalar parameter.}
\label{fig:sag}
\end{algorithm}


\section{Datasets}
\label{s:data}
\subsection{SCAN}
Introduced by~\citet{Lake2018}, SCAN is a collection of tasks used for studying systematic generalization of seq2seq models (see Table~\ref{fig:scan-examples} for some input-output examples). 
A set of 4 primitive verbs are combined with different modifiers generating around 27k unique samples. \citet{Lake2018} and, later, \citet{Loula2018} prepared several non-i.i.d.\ splits of the data into training and test sets. To successfully generalize on such non-i.i.d.\ splits, a model has to generalize systematically, in a compositional way. 

We experiment with three tasks,  often focused on in the literature\footnote{We have also ensured that our best modification performs on par ($\approx$ 10\%) with Transformer on SCAN-length; however SCAN-length is believed to require a different type of generalization~\cite{Gordon2019}.}~\cite{Dessi2019}. \textbf{SCAN-simple} splits all sequences in train and test sets uniformly at random.
Hence, both train and test are identically distributed. Typically models succeed at it easily.
In \textbf{SCAN-jump}, the test set contains all compositional uses of one of the primitives, \texttt{jump}. The train set contains all uses of other primitives, and inputs where \texttt{jump} is used in isolation. 
\textbf{SCAN-around-right} tests if a model is capable to generalize to combinations of two modifiers, \texttt{around} and \texttt{right}, that never co-occur in the training data. The test data contain all  examples where the two modifiers are combined. 

\subsection{Machine Translation}
\label{secsec:MT}
We hypothesize that the type of systematic generalization that SCAN probes for could be most useful in data-poor tasks or tasks with train-test distribution shift. Hence, we complement the standard IWSLT'14 En-De dataset with a low-resource task, FLoRes. To study whether our models can perform SCAN-style generalization on natural language data, we also build a dataset that probes for compositional generalization in noun-adjective ordering in French, when translating from English.

\textbf{IWSLT'14 En-De} 
This is a standard MT dataset, that includes train, validation, \& test sets. We apply preprocessing as in the fairseq example.\footnote{\label{notefairseq}\url{https://github.com/pytorch/fairseq/tree/master/examples/translation}}

\textbf{FLoRes}~\cite{flores} FloRes is a low-resource dataset for English$\leftrightarrow$ Nepali and English $\leftrightarrow$ Sinhala translation. The dataset is split into dev, devtest, and test subsets. We only use the provided supervised data. 

\textbf{Noun-adjective ordering} We take inspiration from SCAN-jump to construct an MT dataset that probes for compositional generalization using noun-adjective ordering in French. 
In French, both \textit{adjective noun} (forward) and \textit{noun adjective} (backward) orders are used, unlike English that only has the forward order.
Which order is used largely depends on the adjective. For example, to refer to a \textit{specific response}, French speakers say \textit{résponse spécifique} (
backward order), while \textit{new response} would be \textit{nouvelle résponse} (forward). 

To draw a parallel with SCAN-jump, we consider the nouns as primitives and adjectives as modifiers. Modifiers appear with different primitives, however, some primitives appear with only  one modifier. For instance, if, in the training set, \textit{response} only appears with \textit{specific} (backward), we test models on translating sentences containing \textit{new response}, where \textit{new} modifies many other nouns in the training set in the forward order.
Such generalization is required by humans 
when dealing with rare or too specific nouns.

To construct our dataset, we start from the English-French Europarl dataset (a part of WMT'14 En-Fr)\footnote{\url{http://www.statmt.org/europarl/}} and select $8$ nouns, $\mathcal{N}=$\{\textit{response, institution, organisation, solution, source, decision, responsibility, population}\}. We constrain our train set so that each of the nouns in $\mathcal{N}$ appears only with one adjective (hence in one particular order) as shown in Table~\ref{tab:compo_dataset}. 
For example, the noun \textit{response} will only be composed with  the adjective  \textit{specific}. However, \textit{specific} (and all other adjectives in Table~\ref{tab:compo_dataset}) appears with other nouns. To select these sentences, we use the CoreNLP parser~\cite{manningetal2014}.\footnote{\url{https://stanfordnlp.github.io/}} 
Finally, all sentences with nouns that are not among the selected ones are kept. In other words, the training  set may contain sentences that have neither the selected adjectives nor the selected nouns.
This results to $1641681$ sentence pairs split into train ($1478195$ pairs) and validation ($163486$ pairs) sets. \\
The test set is composed  of the filtered sentences of the original Europarl dataset: we select sentences where nouns in the backward column of Table~\ref{tab:compo_dataset} (\{\textit{response, institution, organism, solution}\}) are only modified by the adjectives in the forward column (\{\textit{new, good, big, first}\}). Similarly, we also consider the sentences where the nouns of the forward column are composed with the adjectives of the backward column of the Table.\footnote{We use the Stanford parser to select these sentences.} This process will ensure that in the test set, the noun-adjective only appears in the reverse order compared to the train set. Unlike the training data, the test data contains only sentences with the target nouns and adjectives. In total, we test models on $659$ sentences. Note that the train and validation sets are identically distributed, however, the test set is distribution-shifted w.r.t.\ train, akin to SCAN-jump. \\
We follow the preprocessing steps on the fairseq example page for WMT'14 English to French.\textsuperscript{\ref{notefairseq}}\\ 

\begin{table}[]
    \centering
    \resizebox{1\columnwidth}{!}{
    \begin{tabular}{cc}
        backward order  & forward order \\
        \toprule
        \makecell{{(`specific', `response')}\\ {(`particular', `institution')}\\{(`effective', `organisation')} \\  {(`permanent', `solution') }} &
        \makecell{{(`new', `source')} \\  
        {(`good', `decision')}\\      
        {(`big', `responsibility')} \\
        {(`first', `population')}}
 \\
    \bottomrule
    \end{tabular}
    }
    \caption{(adjective, noun) pairs in the train set of the noun-adjective ordering dataset, classified by their order in French language.} \vspace{-.3cm}
    \label{tab:compo_dataset}
\end{table}

\begin{table*}[tb]
    \centering
  \resizebox{0.7\textwidth}{!}{
    \begin{tabular}{lcccccccccccccc}
        & jump & around-right  & 
        simple \\
        \toprule
        Transformer & $3.4_{\pm 2.0}$ & $97.6_{\pm 1.5}$ &  
        $100.0_{\pm 0.0}$ \\
        \midrule
        self-att.\ gate  (SAG)& $17.2_{\pm 5.8}$ & $85.2_{\pm 10.0}$ & 
        $100.0_{\pm 0.0}$ \\
        \quad + Conv as s.-a. & $25.7_{\pm 20.4}$ &  $38.4_{\pm 7.8}$ & 
        $99.8_{\pm 0.0} $\\
        \quad + Fixed-span & $33.6_{\pm 9.5}$ & $97.6_{\pm 1.3}$ & 
        $100.0_{\pm 0.0}$ \\
        \quad + T5 & $43.0_{\pm 9.5}$  & $92.6_{\pm 2.8}$ &  
        $100.0_{\pm 0.0}$ \\
        \midrule
        LSTM seq2seq \cite{Lake2018} & 1.2 &  
        2.5 & 
        99.8 \\
        ConvS2S \cite{Dessi2019} &  $69.2_{\pm 8.2}$ & $56.7_{\pm 10.2}$ &  
        $100.0_{\pm 0.0}$ \\
        \bottomrule
    \end{tabular}
    }
    \caption{Accuracy on SCAN tasks, \%. For each architecture and task, we report the mean accuracy of the best hyperparameter configuration. $\pm$ denotes 1 SEM.}
    \label{tab:scan}
\end{table*}

\section{Methodology}
\label{s:methodology}
\textbf{SCAN} \citet{Lake2018} were concerned by the feasibility of systematic generalization in seq2seq models. Hence, in their experiments, they tuned the models on the train data and then directly evaluated them on test set, reporting test scores. 
We follow the same protocol: given a grid of hyperparameters, we fit models on the training data. Next, for each hyperparameter configuration, we average the performance of the models across random seeds. Such a setup  demonstrates that, at least for some hyperparameter configurations, the introduced models can learn to generalize systematically. At evaluation time, we decode greedily. 

\textbf{IWSLT'14 De-En} We run a grid search on train data; next we select the best performing checkpoint on the validation dataset. We report performance on the test data. We use the same training and evaluation protocols as suggested on the fairseq MT example page.\textsuperscript{\ref{notefairseq}} We use beam size 5. 

\textbf{FLoRes} This dataset has dev, devtest, and test splits provided. We run a hyperparameter grid search training on the dev data. Next, we select the hyperparameter configuration that has the best average (across seeds) performance on devtest. We report the performance of the selected hyperparameter configuration on the test set, averaged across seeds. We use the training/evaluation scripts, tokenization and other parameters suggested on the dataset page: 
beam size 5 and length penalty 1.2.

\textbf{Noun-adjective ordering} We run the hyperparameter search similarly to IWSLT'14 De-En.
The training and evaluation protocols are the ones suggested by the fairseq page for WMT'14 En-Fr.\textsuperscript{\ref{notefairseq}} We also use beam size 5.

As we aim to probe abilities for compositional generalization, we introduce an accuracy-based measure, \emph{COMP}. When analyzing models' errors, we encountered 3 common errors: (1) removing the adjective (example 1 in Table~\ref{tab:qualitative}), (2) replacing the adjective with a synonym and reversing the order (examples 2 and 3 in Table~\ref{tab:qualitative}), and (3) producing a completely wrong generalization while removing the adjective. While (2) provides a good enough translation, it is a mistake in the noun-adjective order. However, when outputting the right noun and adjective, the order is always preserved. Hence, to measure if a model is compositional, we only look if both the target adjective and the target noun appear in the prediction, irrespective of their order. 
We define thus COMP as the ratio of predicted sentences that include \emph{both} the target adjective and noun.\footnote{It happens that models use a synonym in the right order as shown in SAG+T5's prediction 2 in Table~\ref{tab:qualitative}. In that case, models had generalized well but are still penalized by COMP. COMP is hence only a proxy measure for compositional generalization based on the common failures.}

\textbf{Hyperparameter search}
Transformer models have multiple hyperparameters (embeddings dimensionality, number of layers and attention heads, dropout probabilities, etc.). On top of those, our introduced models add the attention span $s$, and the initial gate state $\beta_0$. For MT tasks, we start from the existing strong baseline hyperparameter configurations (FLoRes: specified by~\citet{flores}, De-En \& En-Fr: following the fairseq example page) and only tune (a) the parameters introduced by our architectures, and (b) the attention dropout parameter (for all architectures, including Transformer). 
For SCAN, there is no baseline hyperparameter configuration, so we start with tuning Transformer and then base hyperparameters of the introduced architectures on it.
We report full hyperparameter grids in Supplementary.

\section{SCAN experiments}
\label{s:scan}
In our preliminary experiments, we found that our modifications of the self-attention mechanism do not lead to improvements over the standard Transformer when they are not combined with the self-attention gate (SAG). Hence, we focus our experiments on architectures that include SAG.

We report our results in Table~\ref{tab:scan}. We also include results for LSTM- and Conv-based seq2seq models that were reported in earlier work~\cite{Lake2018,Dessi2019}.
From Table~\ref{tab:scan}, we see that the unmodified Transformer has very low accuracy on \texttt{jump} (3.4\%), which is only slightly above that of LSTM seq2seq (1.2\%) and well below ConvS2S (69.2\%). 
This indicates that Transformer models are indeed failing in compositional generalization on \texttt{jump}. However, they have a very high score on the \texttt{around-right} split (97.6\%) and \texttt{simple} ($\ge$ 99.8\%). 
By introducing the different modifications described in Section~\ref{s:modifications}, making Transformers closer to ConvS2S, we aim at preserving the high performance of Transformers on \texttt{around-right} and \texttt{simple} while significantly improving it on \texttt{jump}.

Adding SAG increases accuracy on \texttt{jump} 5-fold (17.2\%) at the expense of a small drop in \texttt{around-right} scores (not stat.\ sig.).

Further, we observe that changes of the self-attention mechanism (replacing it with Convs, limiting its span, and adding a relative position-dependent bias), can further increase the performance on \texttt{jump}. Apart from SAG+Conv as s.-a, the  self-attention modifications do not significantly alter the performance on \texttt{around-right}. 

We see that the architectural changes that we proposed improve the compositional capabilities of the Transformer models. As expected, the introduced hybrid architectures reach significantly better performance on \texttt{jump} (up to 12x improvements for SAG+T5) while keeping high performance on the \texttt{around-right} \& \texttt{simple} tasks. 



\begin{table}[tb]
    \centering
  \resizebox{1\columnwidth}{!}{
    \begin{tabular}{ccccc}
        Transformer & SAG & + Conv s.-a. & + fix. span &  + T5\\
        \midrule
        $34.64_{\pm 0.03}$ & $34.28_{\pm 0.08}$ & $33.44_{\pm 0.04}$ & $34.32_{\pm 0.01}$ & $34.66_{\pm 0.04}$ \\
        \bottomrule
    \end{tabular}
    }
    \caption{BLEU on test set. IWSLT'14 German to English dataset.  $\pm$ denotes 1 SEM.} \vspace{-.3cm}
    \label{tab:ge-en}
\end{table}

\section{Machine Translation experiments}
\label{s:mt}
\textbf{IWSLT'14 De-En}
In Table~\ref{tab:ge-en}, we report BLEU scores on German-English translation. SAG + T5 performs slightly better (0.02 BLEU, not stat.\ sig.), but other modifications underperform w.r.t.\ Transformer. Replacing self-attention with convolutions resulted in the largest drop, 3\%. Other differences are smaller, $\le$ 1\%. For the following parts, we only experiment with the SAG + T5 model as the only non-degraded one. However, results with the remaining models on FLoRes and the Noun-adjective ordering datasets are reported in Supplementary.

\textbf{FLoRes, Si/Ne $\leftrightarrow$ En}
We report results on English$\leftrightarrow$Nepali and English$\leftrightarrow$Sinhala  translation in Table~\ref{tab:flores-comp}. Following~\citet{flores}, we use tokenized BLEU when translating from English. We run standard Transformer models as specified in \citet{flores}, but adding a search over the attention dropout probability. We verify that we have close results compared to~\citet{flores}.\footnote{We got better BLEU scores due to the extra grid search.}
Table \ref{tab:flores-comp} shows that SAG + T5 outperforms Transformer on all language pairs and directions with relative improvements between 3.6\% (si-en) and 13.1\% (en-ne). 

\begin{table*}[tb]
    \centering
  \resizebox{.85\textwidth}{!}{
    \begin{tabular}{lccccccc}
    \toprule
        & \multicolumn{4}{c}{FLoRes (BLEU)} && \multicolumn{2}{c}{Noun-Adj.~ordering}\\
         \cmidrule{2-5}    \cmidrule{7-8}
        &  ne-en & si-en & en-ne &  en-si && BLEU & COMP\\
        \midrule
        Transformer & $7.94_{\pm 0.05}$ & $7.15_{\pm 0.07}$ & $4.43_{\pm 0.01}$ & $2.32_{\pm 0.08}$ && $40.86_{\pm 0.34}$ & $0.64_{\pm 0.01}$ \\
        SAG + T5 & $\mathbf{8.40_{\pm 0.02}}$ & ${7.41_{\pm 0.10}}$  & $\mathbf{5.01_{\pm 0.10}}$ &  $\mathbf{2.54_{\pm 0.03}}$&& $41.43_{\pm 0.29}$ & $\mathbf{0.73_{\pm 0.01}}$\\
        \bottomrule
    \end{tabular}
    }
    \caption{Models performance on FLoRes and Noun-adjective ordering (English to French) dataset. For FLoRes, we report the BLEU dev-test scores for the different translation directions. For the Noun-adjective ordering dataset, we report both BLEU and COMP measures on the test set. In bold are values that stat. sig. improve over Transformer. $\pm$ denotes 1 SEM.}
    \label{tab:flores-comp}
\end{table*}

\textbf{Noun-adjective ordering}
BLEU scores on the test set are reported in Table~\ref{tab:flores-comp}. SAG + T5 leads to a relative improvement of $1.39\%$ compared  to standard Transformer. 
BLEU, however, is not informative about the particular noun-adjective generalization. We hence also report COMP scores. 
From Table~\ref{tab:flores-comp}, we see that  
SAG + T5 demonstrates a significant improvement with $14\%$ relative gain compared to the standard Transformer architecture. Our follow-up experiments show that the hybrid model recovers an average of 43.3\% of cases where the best Transformer model (best seed w.r.t.\ COMP) failed in compositional generalization, whereas Transformer is only correct at 21.5\% of SAG + T5's errors. We report in Table~\ref{tab:qualitative} examples comparing  SAG + T5 and Transformer translations. 



\textbf{Discussion}
Upon analyzing our experiments on SCAN and machine translation tasks, we see the following picture. Indeed the hybrid models that we described in Section~\ref{s:archs} have considerably higher accuracy on SCAN-\texttt{jump} w.r.t.\ Transformer and a comparable performance on the other SCAN splits. Hence, our results suggest the importance of both gating and (the ability of) limiting the attention span for SCAN generalization. 

As expected, the improvement on SCAN do not consistently entail improvements on the resource-rich dataset, and only the  combination of SAG and T5 showed a tiny improvement. This emphasizes the importance of testing models on realistic setups  to model from being too SCAN-tailored.

Finally, we test SAG + T5 on low-resource and compositional tasks. The hybrid architecture shows consistent improvements on FLoRes for all translation directions, with at up to 13.1\% relative improvement, and on the the natural language compositional task with 14\% relative improvement on COMP. Our qualitative analysis also showed that SAG + T5 correctly handles noun-adjective ordering in most cases, while Transformer makes more mistakes. 

\begin{table*}[!htbp]
    \centering
    \begin{tabular}{p{15cm}}
    \toprule


    \small{\textbf{Target:} Nous sommes face à une \textit{\underline{responsabilité politique particulière}}.}\\

    \small{\textbf{Prediction SAG+T5:}
    Nous sommes accablés par une \textit{\underline{responsabilité politique particulière}}}.\\

    \small{\textbf{Prediction Transformer:} Nous sommes accablés par une \textit{\underline{responsabilité politique}}}.\\
    \midrule
    \midrule
    \small{\textbf{Target:} Nous voulons trouver une \textit{\underline{bonne solution}} à ce problème.}\\
    \small{\textbf{Prediction SAG+T5:} Nous voulons trouver une \textit{\underline{bonne solution}} à ce problème.} \\
     \small{\textbf{Prediction Transformer:} Nous voulons trouver une \textit{\underline{solution adéquate}} à ce problème.} \\
    \midrule
    \midrule
    \small{\textbf{Target:} Ce qui nous déçoit par rapport à cette \textit{\underline{décision particulière}}, c'est que le projet aurait pu clairement voir le jour.}\\
    \small{\textbf{Prediction SAG+T5:} Ce qui est triste dans cette \textit{\underline{décision précise}}, c'est que le projet aurait été clairement réalisé.}\\
    \small{\textbf{Prediction Transformer:}  Ce qui est triste dans cette \textit{\underline{mauvaise décision}}, c'est que le projet aurait clairement été.} \\
    \bottomrule
    \end{tabular}
    \vspace{-.1cm}
    \caption{Generation Examples for Noun-adjective ordering dataset. Models are tested on the  underlined and italic \textit{\underline{(adjective, noun)}}. For the first 2 examples, SAG+T5 predicted the right \textit{\underline{(adjective, noun)}} translation. In the last one, SAG+T5 replaced the adjective with a synonym but in the right target order (the one not seen in the training set). In the first example, Transformer removed the adjective \textit{particulière}. In the two following examples, Transformer replaced the right adjective with a close synonym adjective to be conform with the training order. For instance, in the second example, \textit{bonne} (an adjective that appears in the forward order) was replaced by \textit{adéquate} (an adjective that appears in the backward order)  as the  \textit{solution} appears only in the backward order at training.
    }
    \label{tab:qualitative}
\end{table*}

\section{Related Work}
\label{s:background}

\textbf{Compositionally-biased models} 
Several approaches were proposed to build SCAN-capable architectures. They span from meta-learning~\cite{Lake2019}, disentangling syntax and semantics~\cite{Russin2019}, learning equivariant~\cite{Gordon2019} and disentangled representations~\cite{Li2019} or combining neural \& symbolic computations~\cite{Chen2020}.
In contrast, we do not  build new models that are specialized to SCAN. Instead, we show that a standard model can be incrementally modified so that performs well on SCAN and still performs well on a standard MT task. 
Having such incrementally improved models allows us to step back and wonder if SCAN (or similar artificial tasks) should be used as a guidance when developing new models.

\citet{Bastings2018} raised concerns due to SCAN being too artificial by showing that even degenerate architectures can perform well on some SCAN tasks. Our results echo their findings: by developing architectures tailored for SCAN, one can easily come up with models that perform worse on general tasks. However, we find that if one avoids this ``SCAN overfitting'' and endows a model with capabilities that SCAN probes for without harming its general performance, they can gain in low-resource scenarios and better handle relevant phenomena in language.


\textbf{Changing attention mechanisms} Self- and cross-attention mechanisms were tweaked in earlier work in order to inject useful biases, e.g., by adding information of relative positions of tokens~\cite{Shaw2018,Raffel2019} or accounting for the locality bias in cross-attention~\cite{Yang2018}.  \citet{Sukhbaatar2019} and \citet{Rae2020} demonstrated that having a short attention span on the lower layers of Transformer models is enough for good language modeling performance. 





\section{Conclusion}
\label{s:conclusions}
In this work, we primarily focused on whether and in which scenarios the inductive bias for compositional generalization, that SCAN looks for, can be useful in natural language tasks. 

We ran study in two steps. As the first step, by exploiting ConvS2S/Transformer similarities, we came up with a modification of the Transformer architecture that performs considerably better than vanilla Transformer on SCAN-jump (43.0\% vs 3.4\% accuracy) and performs equally well on SCAN-simple, SCAN-around-right, and on a standard resource-rich MT task (IWSLT'14 De-En).


Next, we tested this modification in low-resource and distribution-shifted setups. 
In the low-resource MT setup (FLoRes Si/Ne$\leftrightarrow$En), we found that our considered architecture improves by up to 13.1\% in BLEU score over the vanilla Transformer. 
Then, we introduced a new dataset that probes specifically for compositional reasoning in natural language. Unlike SCAN, our compositional dataset is built by filtering an existing natural language corpus (EuroParl En-Fr) to probe how models perform noun-adjective ordering under a (minimal) distribution shift. Thus, we are largely closer to testing the compositional generalization required by humans compared to SCAN, and succeeding on the test set requires both compositional reasoning and good language model performance (see examples in Table~\ref{tab:qualitative}). We believe that such a dataset is beneficial for future research to test more complex compositionality skills.
Finally, our experiments on our dataset demonstrated that better SCAN generalization leads to better results on noun-adjective ordering (14\% on COMP).


Our findings indicate the following. Firstly, as hypothesized before~\cite{Dessi2019,Hupkes2018}, the limited attention span provides a useful inductive bias that allows models to perform better on compositional generalization induction, that SCAN probes for.
Further, endowing a model with SCAN-style generalization capabilities \textit{can} lead to improvements in low-resource and distribution-shifted scenarios as long as we ensure that we do not overfit to SCAN. 

We believe that the contribution of diagnostic datasets like SCAN is of great value.  As performance grows on tasks such as MT, identifying gaps where a model's performance lags will become fundamental and will guide us to develop architectures that cover genuine new linguistic grounds and not just overfit to peculiarities of standard datasets. 
%



\section*{Acknowledgments}
The authors are grateful to Marco Baroni and  the reviewers for feedback that helped us to improve our work.

\FloatBarrier
\bibliography{biblio}
\bibliographystyle{acl_natbib}

\newpage

\appendix

\section{Hyperparameter grids}
\label{sec:hyper}
\subsection{SCAN}
For each architecture, we used the same hyperparameter grids for all splits of SCAN.

All models were trained by Adam with default $\beta_1$ and $\beta_2$ parameters,  for 250 epochs, with batch size 256, learning rate $5 \cdot 10^{-4}$, dropout and attention dropout 0.1, random seeds $\{0, 1, 2\}$. We vary the Encoder and Decoder parameters independently: number of attention heads $\{4, 8\}$, embedding dimensions $\{128, 256\}$, FFN dimensions $\{256, 512\}$, and the number of layers $\{4, 6, 8\}$; clip norm 1.0.

For hybrids models, add the following parameters.
\paragraph{SAG} To reduce the search space, we did not vary $\beta_0$, setting it to $-1$.

\paragraph{SAG + CNN as self-attention} Span: $\{2, 4, 6\}$, $\beta_0 = -1$, number of layers: \{4, 6\}.

\paragraph{SAG + fixed span self-attention} Span: $\{2, 4, 6\}$, $\beta_0 = -1$, number of layers: 4.

\paragraph{SAG + T5} Span: $\{2, 4, 6\}$, $\beta_0 = -1$, number of layers: 4.

\subsection{Machine Translation}
\paragraph{De-En}
We start  from the standard architecture suggested by Fairseq examples for IWSLT'14 De-En. That is we share decoder input and output embeddings. Both Encoder and decoder have an embedding size of 512 FFN dimensions of  1024, 4 attention heads, 6 encoder \& decoder layers. We used adam optimizer with learning rate of 5e-4, no clip norm, warm-up learning rate 1e-7, inverse square root learning rate scheduler, 4000 warm-up updates, dropout 0.3, weight decay 1e-4, label smoothing 0.1, max.\ tokens per batch per GPU: 1024, 40 epochs. We used 4 GPUs for  training.

For SAG-enabled architectures, we additionally searched for Encoder's $\beta_0$ in $\{-1, 0\}$ and  $\{1, 0\}$ for Decoder. We varied attention span in $\{2, 4, 6\}$, ttention dropout in $\{0.0, 0.2\}$ and pre-block encoder and decoder normalization.

For model selection, we also follow Fairseq example and checkpoint the best model on validation set based on BLEU score. BLEU score is computed with a beam of size 5.

\paragraph{FLoRes}
We used shared embeddings between Encoder and Decoder, embedding dimenions of 512, FFN dimensions of 2048, 2 attention heads, 5 encoder \& decoder layers. We used pre-block normalization\footnote{\texttt{--encoder-normalize-before} and \texttt{--decoder-normalize-before} in fairseq.}, learning rate of 1e-3, no clip norm, warm-up learning rate 1e-7, inverse square root learning rate scheduler, 4000 warm-up updates, dropout 0.4, activation dropout 0.2, weight decay 1e-4, label smoothing 0.2, max.\ tokens per batch per GPU: 4000, 100 epochs. We searched for attention dropout in $\{0.0, 0.2\}$. We used 4 GPUs for training.

For SAG-enabled arcchitectures, we additionally searched for Encoder's $\beta_0$ in $\{-2, -1, 0\}$ and  $\{2, 1, 0\}$ for Decoder. We varied attention span in $\{2, 4, 6\}$.

\paragraph{Noun-adjective order agreement}

We start  from the standard architecture suggested  by Fairseq examples for WMT'14 En-Fr. That is we  share encoder, decoder and output embeddings. Both Encoder and decoder have an embedding size of 1024 FFN dimensions of  4096, 16 attention heads, 6 encoder \& decoder layers. We used adam optimizer with learning rate of 7e-4, no clip norm, warm-up learning rate 1e-7, inverse square root learning rate scheduler, 4000 warm-up updates, dropout 0.1, label smoothing 0.1, max.\ tokens per batch per GPU: 4000, 30 epochs. We used 6 GPUs for  training.

For SAG-enabled architectures, we additionally searched for Encoder's $\beta_0$ in $\{-1, 0\}$ and  $\{1, 0\}$ for Decoder. We varied attention span in $\{2, 4, 6\}$, ttention dropout in $\{0.0, 0.2\}$ and pre-block encoder and decoder normalization.

Best checkpoint is based on the loss of the  validation set.

\section{Other modifications on FloRes and Noun-adjective ordering datasets}
In the main paper, we only experimented with SAG + T5 as the only non-degraded modification on the IWSLT’14 En-De dataset. Our intuition is that the remaining hybrid models are SCAN-tailored and would not lead to any improvement in the low-resource (FloRes) and domain-shifted (Noun-adjective ordering dataset) settings. In this section, we verify our intuition and report the results of all the introduced variants. The hyper-parameters search is reported in Section~\ref{sec:hyper}.

\paragraph{FloRes}
We report results on English$\leftrightarrow$Nepali and English$\leftrightarrow$Sinhala  translation in Table~\ref{tab:flores}. We also report \citet{flores} results under ``Baseline''.

Analyzing results of SAG, we notice that it is usually very close to Transformer's results on all tasks, apart from Nepali$\rightarrow$English, where it lags behind. The fixed-span modification performs worse than Transformer on in all directions. Replacing self-attention with convolutions results in better scores on En$\rightarrow$Ne and worse scores on Ne$\rightarrow$En/En$\rightarrow$Si. 

Hence, as expected, only the SAG + T5 model outperforms Transformer on all language pairs and directions, highlighting the importance of verifying the generality of the model on realistic datasets.

\begin{table*}[tb]
    \centering
  \resizebox{0.7\textwidth}{!}{
    \begin{tabular}{lcccccccccccccc}
        & Baseline & Transformer & SAG & + Conv s.-a. & + fixed span & + T5 \\
        \midrule
        ne-en & 7.6 & $7.94_{\pm 0.05}$ & $7.58_{\pm 0.06}$  & $7.59_{\pm 0.02}$ &   $7.44_{\pm 0.08}$ & $\mathbf{8.40_{\pm 0.02}}$  \\
        si-en & 7.2  & $7.15_{\pm 0.07}$ & $7.14_{\pm 0.10}$ & $7.18_{\pm 0.10}$ & $.78_{\pm 0.07}$
        & ${7.41_{\pm 0.10}}$\\
        en-ne & 4.3 & $4.43_{\pm 0.01}$ & $4.36_{\pm 0.08}$ & $\mathbf{4.63_{\pm 0.03}}$ & $4.12_{\pm 0.05}$ & $\mathbf{5.01_{\pm 0.10}}$\\
        en-si & 1.2 & $2.32_{\pm 0.08}$ & $2.37_{\pm 0.10}$ & $2.14_{\pm 0.03}$ &  $2.12_{\pm 0.04}$ & $\mathbf{2.54_{\pm 0.03}}$\\
        \bottomrule
    \end{tabular}
    }
    \caption{BLEU dev-test scores on FLoRes. Baseline scores are taken from~\cite{flores}. In bold are values that stat.\ sig.\ improve over Transformer ($p < 10^{-3}$).  $\pm$ indicates 1 SEM.}
    \label{tab:flores}
\end{table*}

\paragraph{Noun-adjective order agreement}

BLEU scores on the test set are reported in Table~\ref{tab:en-fr}. SAG leads to a relative improvement of $1.44\%$ compared  to standard Transformer, closely followed by SAG + T5. Still, in total, the differences are very small across all models. On the other hand, all introduced variants outperform standard Transformer on COMP. However, only SAG + T5 demonstrates a significant improvement with $14\%$ relative gain. 

Overall, we observe that the SCAN-tailored variants do not degrade performances on the Noun-adjective order agreement dataset, but still do not lead to any significant improvement,  contrary to SAG + T5.

\begin{table*}[tb]
    \centering
  \resizebox{0.7\textwidth}{!}{
    \begin{tabular}{lccccc}
        & Transformer & SAG & + Conv s.-a. & + fixed span & + T5 \\
        \midrule        
        BLEU & $40.86_{\pm 0.34}$  & $41.45_{\pm 0.14}$ & $39.89_{\pm 0.27}$ & $41.01_{\pm 0.24}$ & $41.43_{\pm 0.29}$\\
        COMP & $0.64_{\pm 0.01}$ & $0.70_{\pm 0.03}$ & $0.67_{\pm 0.01}$ & $0.68_{\pm 0.01}$ & $\mathbf{0.73_{\pm 0.01}}$\\
        \bottomrule
    \end{tabular}
    }
\caption{BLEU and COMP measures on test sets: compositional English to French dataset. In bold are values that stat. sig. improve over Transformer ($p < 0.05)$. $\pm$ denotes 1 SEM.}
    \label{tab:en-fr}
\end{table*}

\section{Visualizing attention biases}
In this Section, we illustrate how a successful SAG + T5 model uses  its $b_{ij}$ terms (Eq.~1) to control its attention. 

\begin{figure*}
\centering
\begin{subfigure}{0.25\textwidth}
\centering
  \includegraphics[width=1\linewidth]{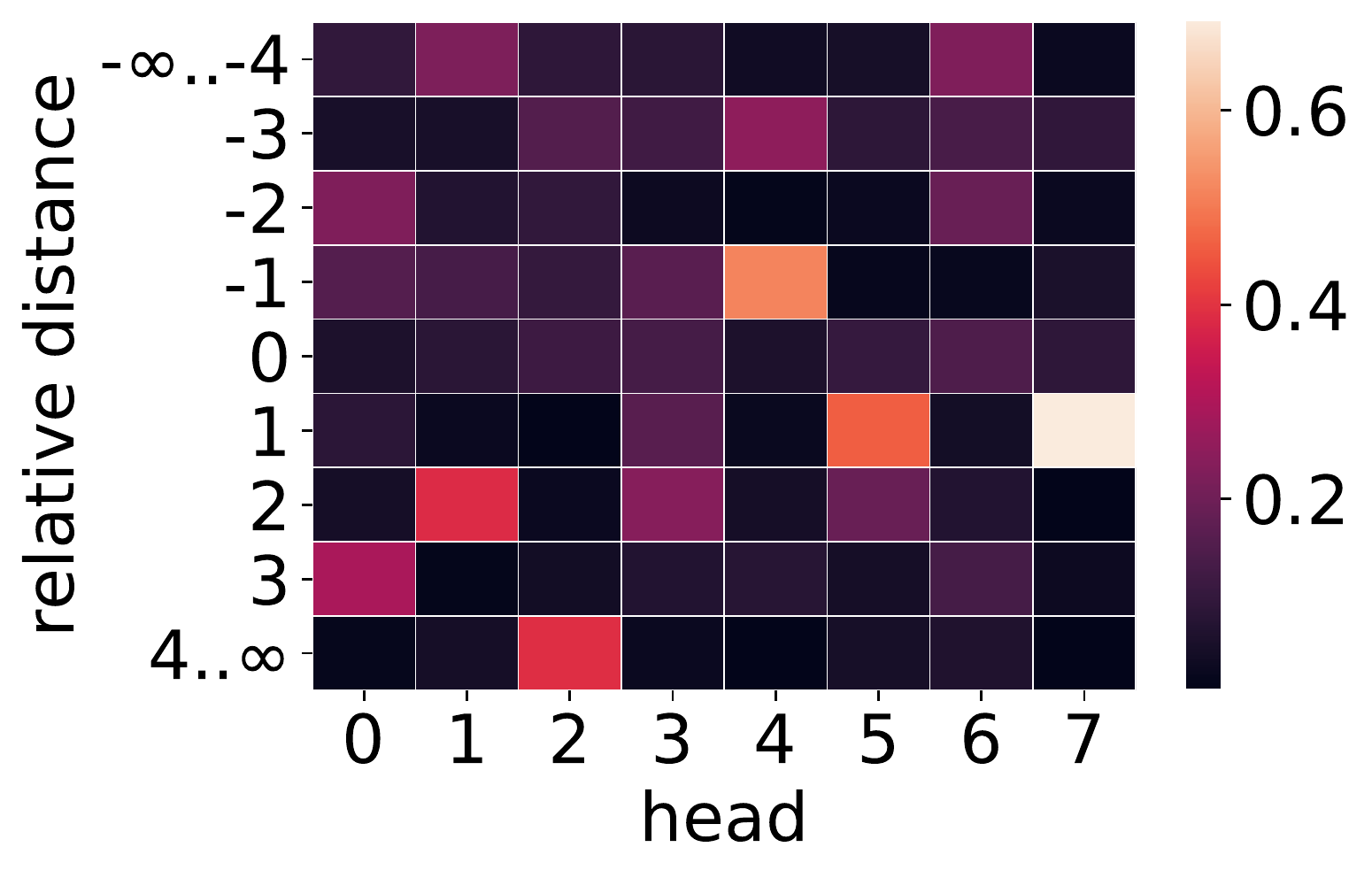}
\caption{Encoder layer 0.}
 \end{subfigure}%
\begin{subfigure}{0.25\linewidth}
\centering
  \includegraphics[width=1\linewidth]{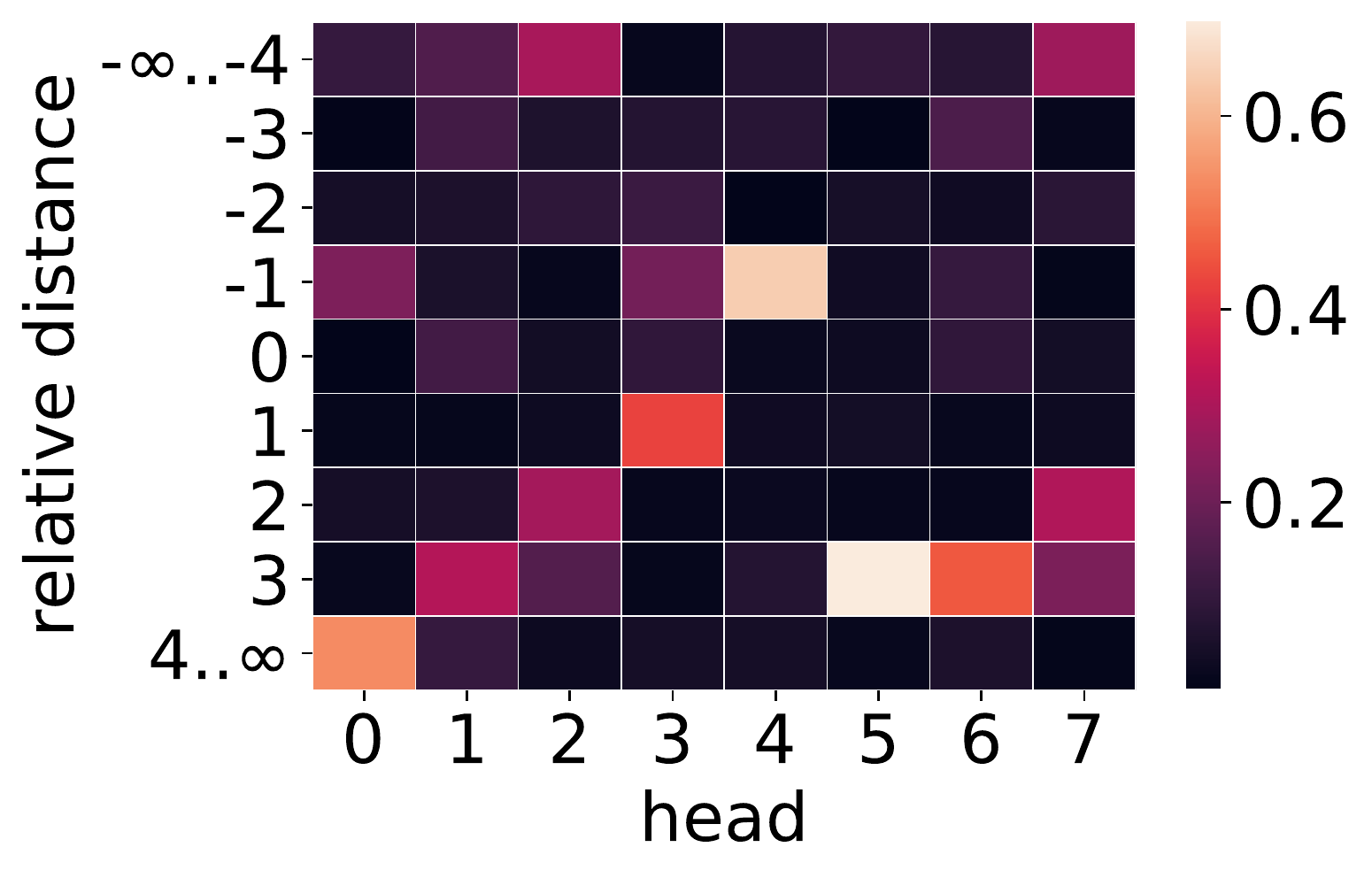}
\caption{Encoder layer 1.}
 \end{subfigure}%
\begin{subfigure}{0.25\linewidth}
\centering
  \includegraphics[width=1\linewidth]{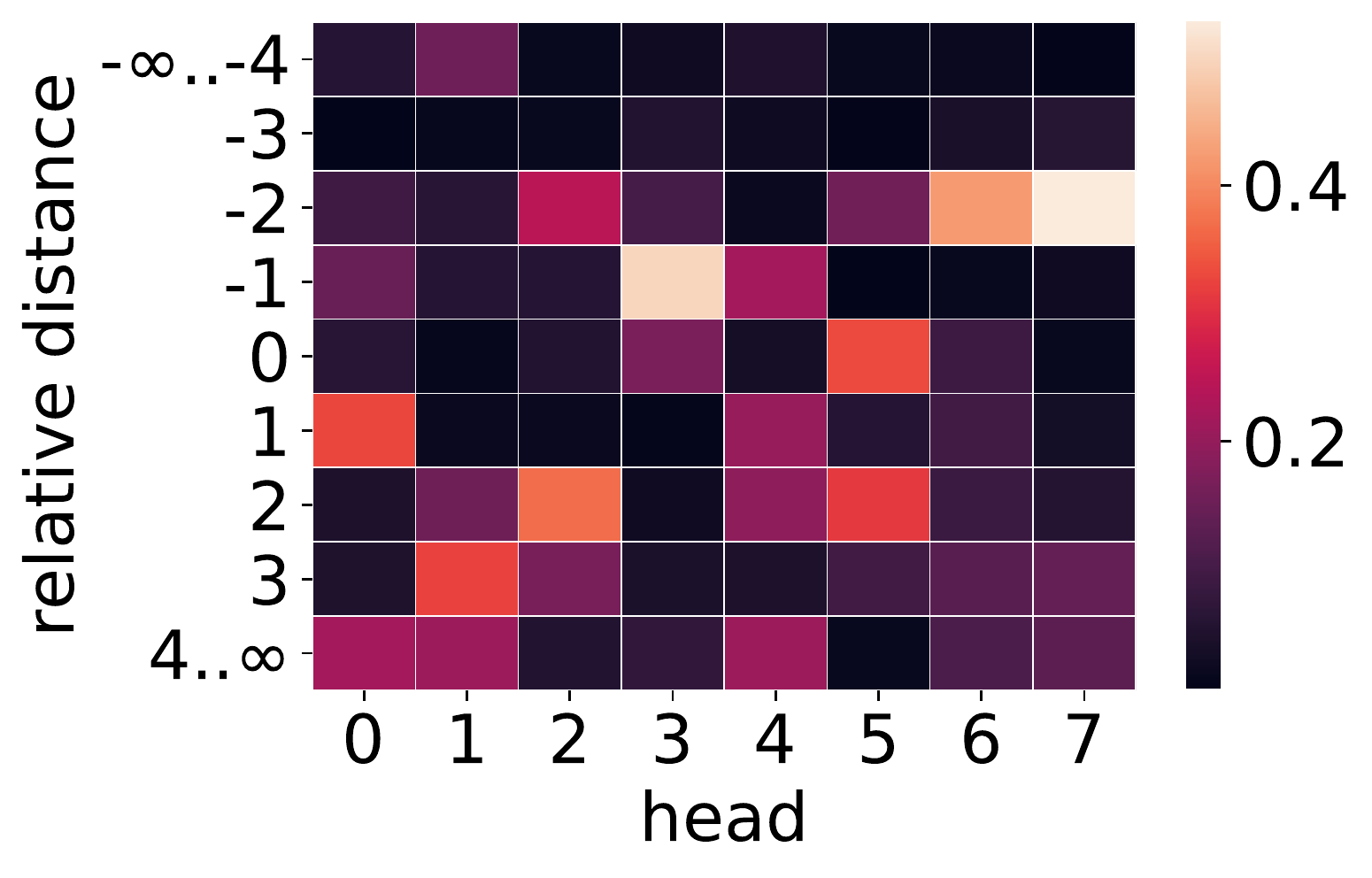}
\caption{Encoder layer 2.}
 \end{subfigure}%
 \begin{subfigure}{0.25\linewidth}
\centering
  \includegraphics[width=1\linewidth]{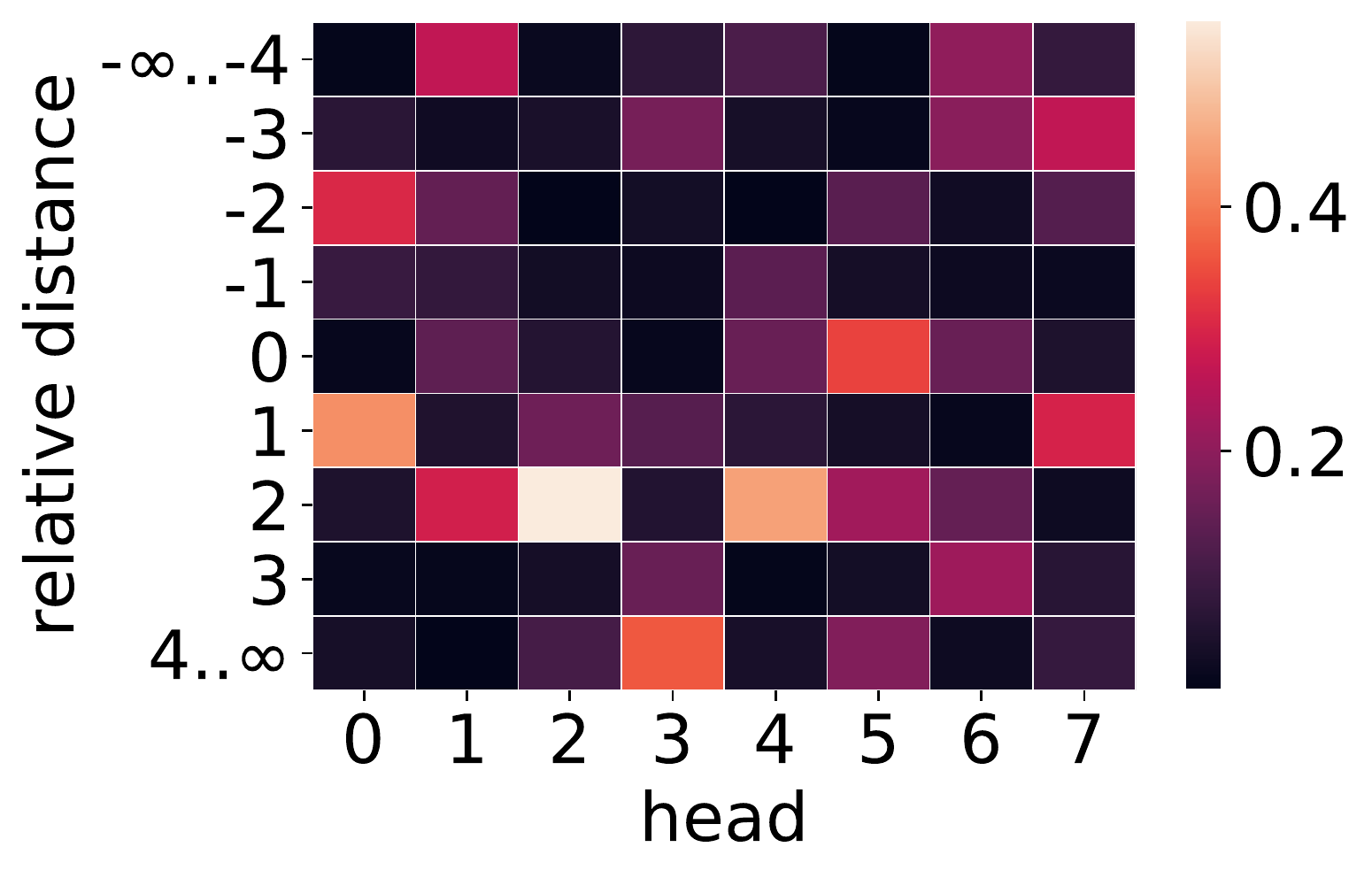}
\caption{Encoder layer 3.}
 \end{subfigure}
\begin{subfigure}{0.25\textwidth}
\centering
  \includegraphics[width=1\linewidth]{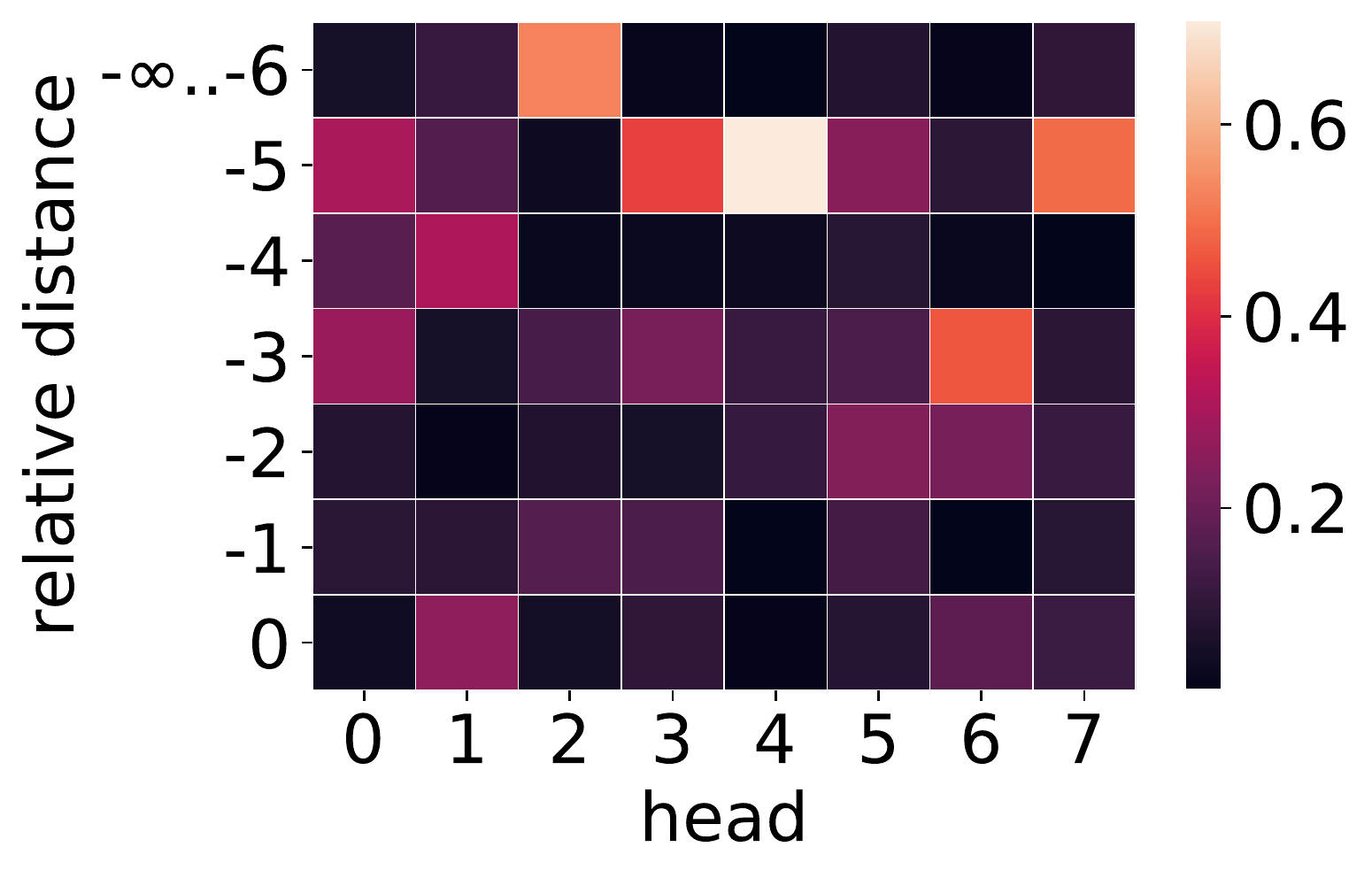}
\caption{Decoder layer 0.}
 \end{subfigure}%
\begin{subfigure}{0.25\linewidth}
\centering
  \includegraphics[width=1\linewidth]{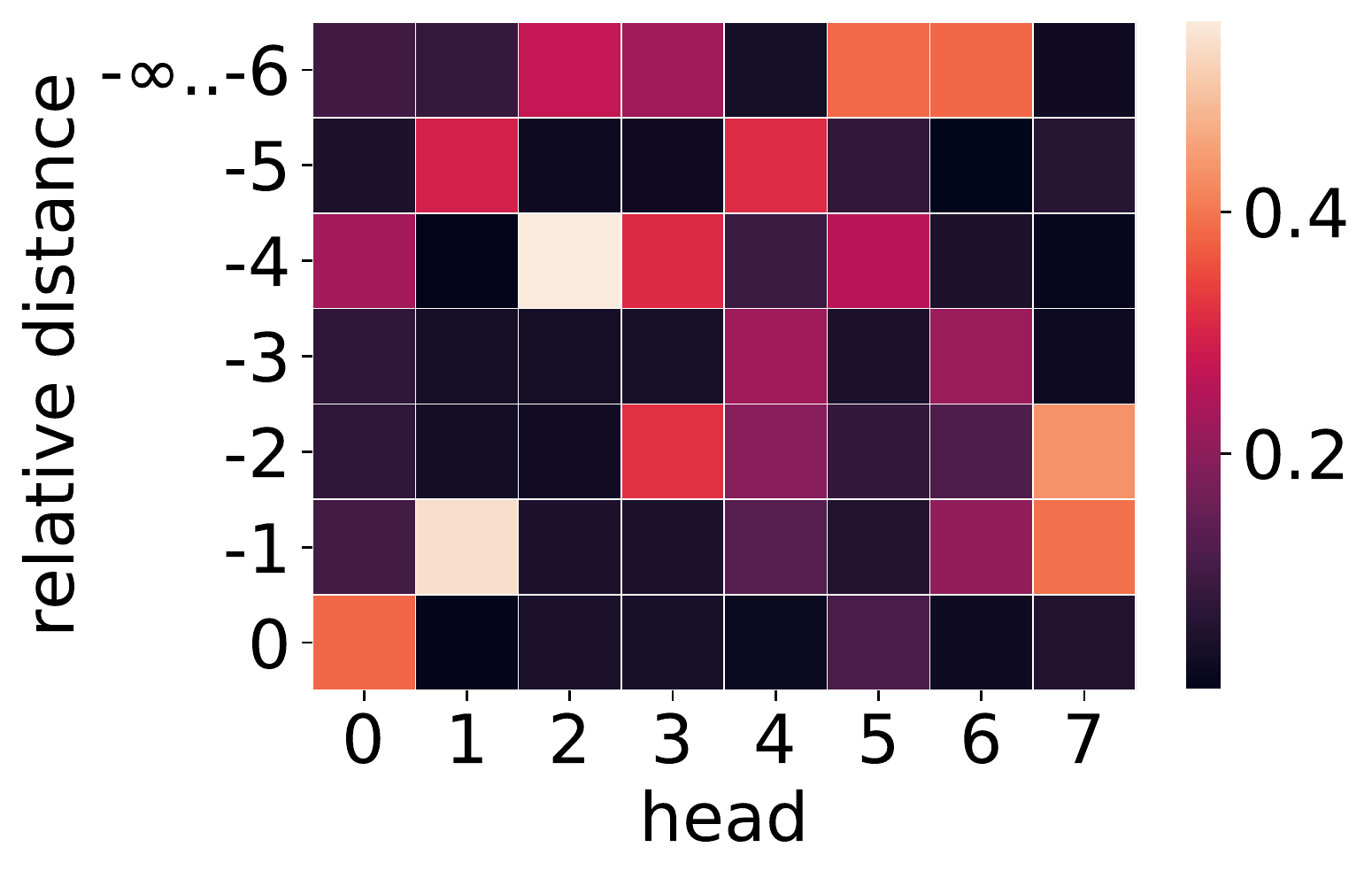}
\caption{Decoder layer 1.}
 \end{subfigure}%
\begin{subfigure}{0.25\linewidth}
\centering
  \includegraphics[width=1\linewidth]{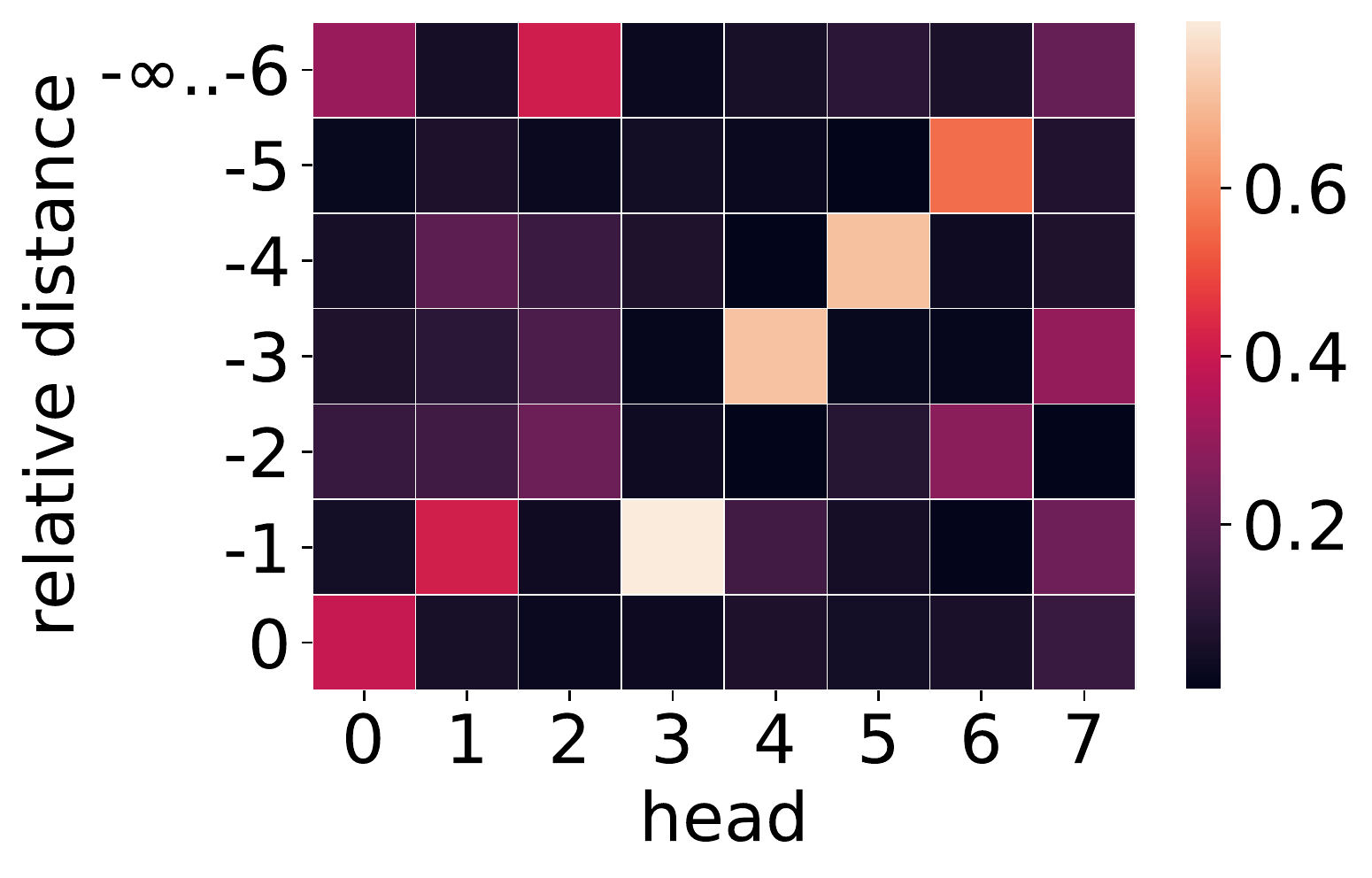}
\caption{Decoder layer 2.}
 \end{subfigure}%
 \begin{subfigure}{0.25\linewidth}
\centering
  \includegraphics[width=1\linewidth]{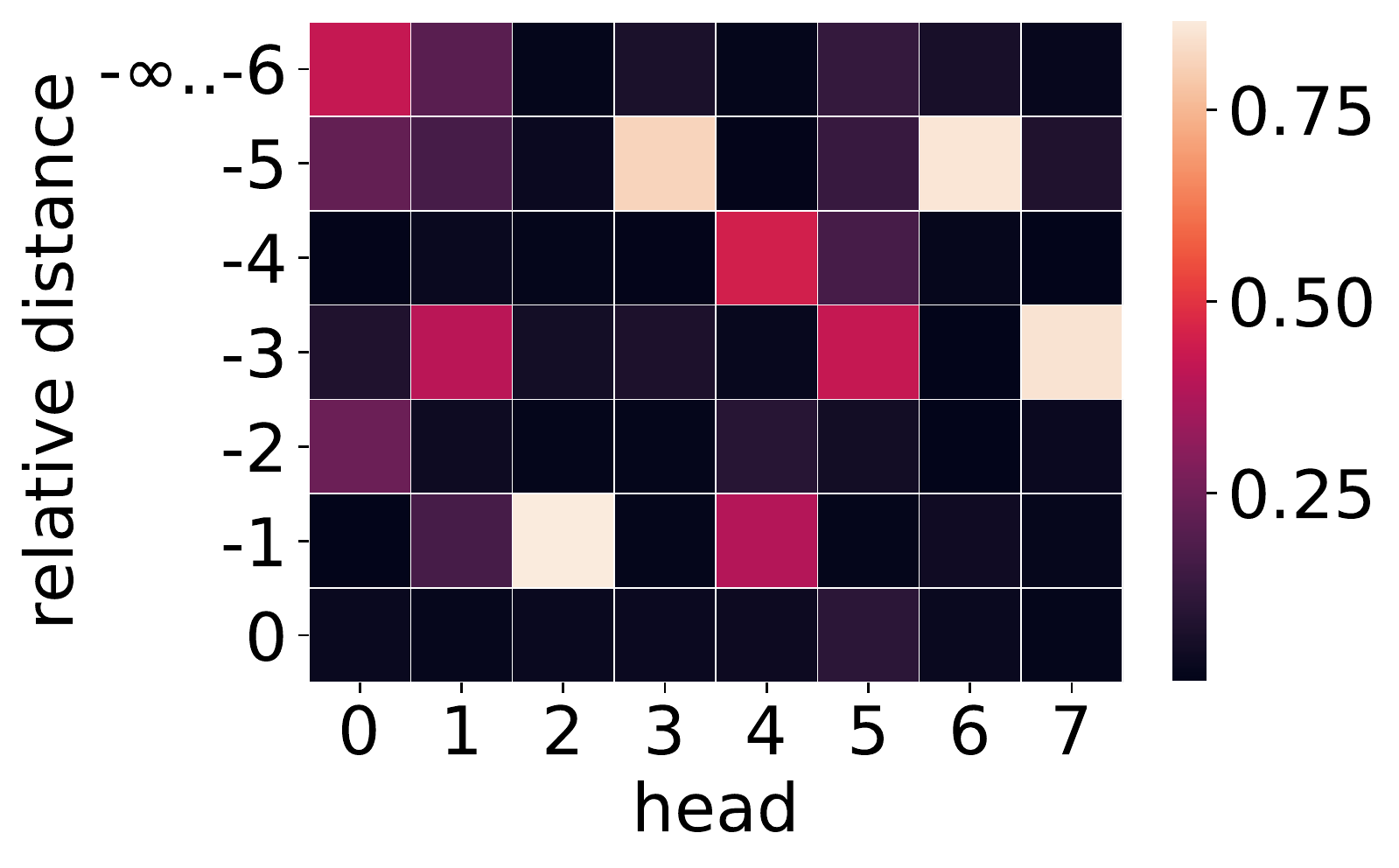}
\caption{Decoder layer 3.}
 \end{subfigure}
 
\caption{Relative attention biases for T5 + SAG architecture (after a softmax). Each cell indicates preference of a head to a position at a signed relative distance. The relative distances are capped. For the decoder we only represent relative attention biases for $d\leq 0$, as  positions with positive relative distance are masked in the autoregressive decoder.}
\label{fig:bij}
\end{figure*}

We take the most successful hyperparameter combination on SCAN-\texttt{jump} in Table~3 and select a model instance that has the best accuracy ($\approx$ 60\%). Next, for each attention head of each encoder and decoder layer, we retrieve its learned relative-position bias $b_d$, where $d$ is a (signed) relative distance between positions $i$ and $j$, that is capped to be within $[-s, +s]$ (see Section~3). For each head, we apply a softmax, to find its ``preference'' $\hat b_d$ over relative positions $d$:
$$
\hat b_d = \frac{\exp(b_d)}{\sum_d \exp(b_d)}
$$
We report the results in Figure~\ref{fig:bij}. Interestingly, quite a few attention heads have very strong preferences for fixed relative positions and some are even dominantly focused on particular positions (Encoder: head 7 in the layer 0; heads 4, 5 in layer 1, heads 3, 7 in layer 2; head 2 in layer 3; Decoder: head 4 in layer 0, head 2 in layer 1, heads 3,4,5 in layer 2; heads 2, 6, 7 in layer 3)\footnote{T5 reduces to the vanilla Transformer if all $b_d$ are equal to zero. That corresponds to the uniform bias $\hat b_d$.}. More often than not, those ``specialized'' heads look within the span and not on the ``border'' values of $d$ (due to $d$ being capped, they also correspond to arbitrary distant positions to the left and right). 

Hence we conclude that in a T5 model (among most successful on SCAN-\texttt{jump}), several heads leverage the ability pay attention locally; supporting our finding that local attention is connected  with the compositional generalization needed to succeed at SCAN. At the same time, some heads have large relative-position bias for distant positions ($[s, +\infty[$ or $]-\infty, -s]$). This general ability to \textit{optionally} look beyond a fixed span in T5 could be responsible for its better performance compared to the fixed span modification.

\end{document}